\documentclass[letterpaper, 10 pt, conference]{ieeeconf}  %

\IEEEoverridecommandlockouts                              %

\usepackage{graphicx} %
\usepackage{times} %
\usepackage{amsmath} %
\usepackage{amssymb}  %
\usepackage{siunitx} 
\usepackage{tikz}
\usepackage[ruled,vlined,noend, linesnumbered]{algorithm2e}
\usepackage{cite}
\usepackage{hyperref}
\usepackage[capitalise]{cleveref}

\usepackage{pgfplots}
\usepgfplotslibrary{colorbrewer}
\pgfplotsset{
	compat = 1.15, 
	cycle list/Set1-8,
	yticklabel style={
		/pgf/number format/fixed,
		/pgf/number format/precision=5
	},
	scaled y ticks=false
}
\usetikzlibrary{pgfplots.statistics, pgfplots.colorbrewer}
\usepackage{pgfplotstable}
\usepackage{standalone}
\usepackage[caption=false]{subfig}
\usepackage{booktabs}
\usepackage{printlen} %

\DeclareMathOperator{\sign}{sign}
\DeclareMathOperator{\ball}{ball}
\DeclareMathOperator{\sd}{sd}

\DeclareMathOperator{\atan2}{atan2}
\DeclareMathOperator{\clip}{clip}
\newcommand{\tb}{\ensuremath{t_{S}}}
\newcommand{\tv}{\ensuremath{t_{V}}}
\newcommand{\td}{\ensuremath{t_{D}}}
\newcommand{\dt}{\ensuremath{\Delta t}}
\newcommand{\ub}{\ensuremath{\boldsymbol{u}}}
\newcommand{\xb}{\ensuremath{\boldsymbol{x}}}
\newcommand{\wb}{\ensuremath{\boldsymbol{w}}}
\newcommand{\Us}{\ensuremath{\mathcal{U}}}
\newcommand{\Xs}{\ensuremath{\mathcal{X}}}
\newcommand{\Ws}{\ensuremath{\mathcal{W}}}
\newcommand{\chix}[1]{\ensuremath{\chi_{\text{#1}}}}
\newcommand{\phix}[1]{\ensuremath{\Phi_{\text{#1}}}}
\newcommand{\uphix}[1]{\ensuremath{\ub_{\text{#1}}}}

\newcommand{\RX}[1]{\ensuremath{\mathcal{R}_{#1}}}
\newcommand{\RS}{\ensuremath{\RX{S}}}

\newcommand{\RZ}{\ensuremath{\RX{Z}}}
\newcommand{\obsj}[1]{\ensuremath{{#1}^{j}}}
\newcommand{\Robsj}{\ensuremath{\obsj{\mathcal{R}}}}

\newcommand{\xm}{\ensuremath{\mathrm{x}}}
\newcommand{\ym}{\ensuremath{\mathrm{y}}}
\newcommand{\zm}{\ensuremath{\mathrm{z}}}
\newcommand{\uamax}{\ensuremath{u_{1, \text{max}}}}
\newcommand{\ubmax}{\ensuremath{u_{2, \text{max}}}}
\newcommand{\pb}{\ensuremath{\boldsymbol{p}}}
\newcommand{\vb}{\ensuremath{\boldsymbol{v}}}
\newcommand{\obs}[1]{\ensuremath{{#1}_{\text{obs}}}}

\newcommand{\rv}{\ensuremath{r_{V}}}
\newcommand{\Oobs}{\ensuremath{\obs{\mathcal{O}}}}
\newcommand{\Oexp}{\ensuremath{\mathcal{O}_{\text{exp}}}}
\newcommand{\Oexpj}{\ensuremath{\mathcal{O}_{\text{exp}, j}}}
\newcommand{\pnull}{\ensuremath{\pb_0}}
\newcommand{\pv}{\ensuremath{\pb_{V}}}
\newcommand{\pL}{\ensuremath{\pb_{L}}}
\newcommand{\Bhat}{\ensuremath{\mathcal{B}}}

\crefname{section}{Sec.}{Sections}
\Crefname{section}{Sec.}{Sections}

\definecolor{TUMblue}{rgb}{0.00, 0.40, 0.74}
\definecolor{TUMdarkblue}{rgb}{0.00, 0.32, 0.576}
\definecolor{TUMlightblue}{rgb}{0.392, 0.627, 0.784}
\definecolor{TUMlighterblue}{rgb}{0.596, 0.776, 0.917}
\definecolor{TUMgray}{rgb}{0.6, 0.6, 0.6}
\definecolor{TUMlightgray}{rgb}{0.855, 0.843, 0.796}
\definecolor{TUMorange}{rgb}{0.89, 0.447, 0.133}
\definecolor{TUMgreen}{rgb}{0.635, 0.678, 0.00}
\definecolor{mygreen}{RGB}{0, 146, 0}
\definecolorset{RGB}{PLOT}{}{%
	Brown, 153, 79, 0;%
	Black,   0,   0,   0;%
	Blue, 86, 180, 233;%
	DarkBlue, 0, 101, 189;%
	Orange, 230, 159, 0;%
	DarkOrange, 213, 94, 0;%
	Purple, 93, 58, 155;%
	Pink, 220, 38, 127;%
	Red, 227, 27, 35;%
	Yellow, 255, 195, 37;%
	BluishGreen, 0, 158, 115;%
	ReddishPurple, 204, 121, 167;%
	Olive, 162, 173, 0;%
	Gray, 153, 153, 153;%
	Cyan, 64, 176, 166;%
	DarkPurple, 136, 34, 85
}

\usetikzlibrary{arrows}
\usetikzlibrary{positioning}
\tikzstyle{gray_box}=[draw, fill=TUMgray, minimum size=2em]
\tikzstyle{blue_box}=[draw, fill=TUMblue, text=white, minimum size=2em]

\SetCommentSty{mycommfont}

\title{\LARGE \bf
Reducing Safety Interventions in Provably Safe Reinforcement Learning
}

\author{Jakob Thumm$^{1}$, Guillaume Pelat$^{1}$, and Matthias Althoff$^{1}$%
\thanks{$^{1}$The authors are with the School of Informatics, Technical University of Munich, 85748 Garching, Germany.
        {\tt\small \{jakob.thumm, guillaume.pelat, althoff\}@tum.de}}%
\thanks{\textcopyright~2023 IEEE.  Personal use of this material is permitted.  Permission from IEEE must be obtained for all other uses, in any current or future media, including reprinting/republishing this material for advertising or promotional purposes, creating new collective works, for resale or redistribution to servers or lists, or reuse of any copyrighted component of this work in other works.}
}

\begin{document}

\maketitle
\thispagestyle{empty}
\pagestyle{empty}

\begin{abstract}
Deep Reinforcement Learning (RL) has shown promise in addressing complex robotic challenges.
In real-world applications, RL is often accompanied by failsafe controllers as a last resort to avoid catastrophic events.
While necessary for safety, these interventions can result in undesirable behaviors, such as abrupt braking or aggressive steering.
This paper proposes two safety intervention reduction methods: proactive replacement and proactive projection, which change the action of the agent if it leads to a potential failsafe intervention.
These approaches are compared to state-of-the-art constrained RL on the OpenAI safety gym benchmark and a human-robot collaboration task.
Our study demonstrates that the combination of our method with provably safe RL leads to high-performing policies with zero safety violations and a low number of failsafe interventions.
Our versatile method can be applied to a wide range of real-world robotic tasks, while effectively improving safety without sacrificing task performance.
\end{abstract}

\section{INTRODUCTION} \label{Sec_Intro}
    \subsection{Motivation}
Reinforcement learning (RL) has emerged as a promising approach for solving complex tasks in a variety of robotic applications, such as manipulation~\cite{lillicrap_2016_ContinuousControl, andrychowicz_2020_LearningDexterous, liu_2021_DeepReinforcementa}, mobile robots~\cite{tai_2017_VirtualtorealDeep, zhu_2021_DeepReinforcement, chang_2021_ReinforcementBased}, and autonomous driving~\cite{elsallab_2017_DeepReinforcement, wang_2021_CommonRoadRLConfigurable, kiran_2022_DeepReinforcement}.
Recent works have shown that safety can be guaranteed for RL agents using additional system knowledge~\cite{berkenkamp_2017_SafeModelBased, alshiekh_2018_SafeReinforcement, fulton_2018_SafeReinforcement, pham_2018_OptLayerPractical, cheng_2019_EndtoEndSafe, krasowski_2020_SafeReinforcement, shao_2021_ReachabilitybasedTrajectory, hunt_2021_VerifiablySafe, thumm_2022_ProvablySafe}.
However, these safety mechanisms can result in undesirable behavior, such as sudden braking or high torques, which can be uncomfortable for humans or demanding for robot hardware.
In addition, bringing the system to an invariably safe state (ISS) \cite{liu_2020_ProvablySafeCooperative}, e.g., a full stop through a failsafe maneuver, can lead to significant recovery times.
Consequently, keeping the number of safety interventions as low as possible is desirable.

This work addresses the problem of safe reach-avoid robotic tasks in environments with static and dynamic obstacles.
Specifically, the robot aims to reach a target position while avoiding collisions with obstacles.
In this work, we guarantee the safety of robots using a generalized version of our proposed shield in~\cite{thumm_2022_ProvablySafe}.
We seek to minimize the number of safety interventions while guaranteeing safety to promote natural and interference-free robot behavior.

\begin{figure}[t]
    \vspace{0.17cm} 
    \centering
    \includegraphics[scale=0.95]{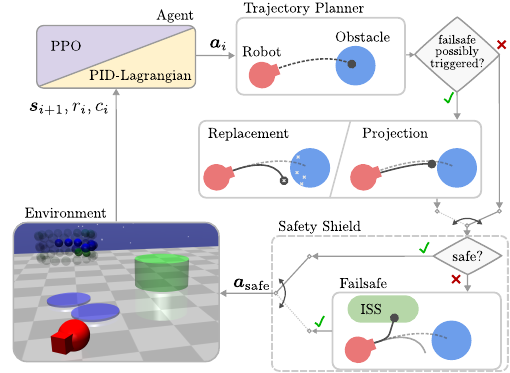}
    \caption{Proposed intervention reduction technique:
    First, the agent generates an action $a_i$, which is then transformed into an intended trajectory via a trajectory planner.
    In case the intended trajectory would potentially trigger a failsafe controller, we replace it using proactive replacement or proactive projection.
    We verify the chosen trajectory during execution and fall back on a failsafe trajectory that ends in an invariably safe state (ISS) when needed.
    }
    \label{fig:main_idea}
\end{figure}

Our two proposed approaches for achieving the desired reduction of safety interventions are \emph{proactive replacement} and \emph{proactive projection}.
The integration of these methods in the safe RL learning cycle is illustrated in \cref{fig:main_idea}.
We first check if the RL action potentially triggers a failsafe intervention.
If that is the case, we can either proactively replace the action with a verified randomized action, or proactively project the action to a verified action that lies nearby in Euclidean space.
Our safety shield and the proposed proactive replacement and projection methods utilize set-based reachability analysis, making them real-time capable for most high-dimensional systems.
We incorporate all possible obstacle trajectories with limited prior knowledge, system and measurement uncertainties, and time delays in our reachability analysis.
To our best knowledge, we are the first to propose a technique that can significantly reduce the number of safety interventions in provably safe RL for any robotic environment.
We compare our two approaches to constrained RL, which tries to solve the original RL problem while keeping the number of safety interventions below a predetermined threshold.

\subsection{Related work}
The safety of RL agents has been a significant concern since the introduction of RL and deep RL~\cite{altman_1998_ConstrainedMarkov, garcia_2015_ComprehensiveSurvey}.
To broadly categorize safe RL, Brunke et al.~\cite{brunke_2022_SafeLearning} classify safe RL methods according to their types of constraints, including soft, probabilistic, and hard constraints.

This work does not rely on probabilistic constraint methods, so readers interested in this topic are referred to \cite{brunke_2022_SafeLearning}.
The most common soft constraint method is constrained RL, which aims to learn the policy with the highest reward while limiting the number of constraint violations~\cite{altman_1998_ConstrainedMarkov}.
The Lagrangian method~\cite{altman_1998_ConstrainedMarkov} is a popular way to address constrained RL problems by converting them to a dual problem, following constrained optimization theory~\cite[Chapter~5]{boyd_2004_ConvexOptimization} and optimizing the Lagrangian multiplier in conjunction with the RL policy.
More recent works, such as constrained policy optimization \cite{achiam_2017_ConstrainedPolicy}, constrained RL with a proportional–integral–derivative-controlled Lagrange multiplier (PID-Lagrangian) \cite{stooke_2020_ResponsiveSafety}, and worst-case soft actor-critic \cite{yang_2021_WCSACWorstcase} build on the Lagrangian method and make it applicable to deep RL.
A drawback of constrained RL is that it cannot guarantee safety in a provable way, as the learned behavior is not formalized or proven.
Additionally, most constrained RL implementations have a non-zero constraint violation threshold, making it impossible to ensure safety with complete certainty.

This work falls in the group of provably safe RL, which fulfills hard constraints by incorporating additional system knowledge in the form of abstract models into the RL framework.
Krasowski et al.~\cite{krasowski_2022_ProvablySafe} categorizes provably safe RL into three classes based on the methods used to alter the actions of the RL agent: action replacement \cite{fisac_2019_GeneralSafety, alshiekh_2018_SafeReinforcement, shao_2021_ReachabilitybasedTrajectory, bastani_2021_SafeReinforcement}, action projection \cite{pham_2018_OptLayerPractical, cheng_2019_EndtoEndSafe, gros_2020_SafeReinforcement, wabersich_2021_PredictiveSafety}, and action masking \cite{krasowski_2020_SafeReinforcement, fulton_2018_SafeReinforcement, mason_2017_AssuredReinforcement}.
However, both action projection and replacement can cause noticeable interventions with the adverse effects mentioned earlier.
To address this issue, some works have used a negative reward to penalize safety interventions~\cite{alshiekh_2018_SafeReinforcement, pham_2018_OptLayerPractical, shao_2021_ReachabilitybasedTrajectory}.
Nonetheless, this approach requires careful hand-tuning and often results in either no reduction of interventions or significant performance loss, as we will discuss in more detail later.
To evaluate the effectiveness of safe RL methods, the OpenAI safety gym \cite{ray_2019_BenchmarkingSafe} has become a widely used benchmark, offering a range of tasks that require balancing performance and safety considerations.

\subsection{Contributions}
Our contributions are fourfold.
First, we introduce the first RL agent to the OpenAI safety gym~\cite{ray_2019_BenchmarkingSafe} that guarantees zero constraint violations. 
Second, we propose proactive replacement and proactive projection, two particular forms of action replacement and projection, that significantly reduce the number of failsafe interventions in safe RL and compare it to constrained RL.
Third, we evaluate our proposed approaches in the OpenAI safety gym and a human-robot collaboration task\footnote{Our code and experimental evaluation are available at \href{https://github.com/JakobThumm/safety-intervention-reduction}{https://github.com/JakobThumm/safety-intervention-reduction}.}.
Finally, we demonstrate our proposed method's real-world impact using a six degree-of-freedom (DoF) manipulator in a human environment.

\subsection{Article structure}
\cref{Sec_Preliminaries} introduces the necessary notation for the RL approaches and set-based reachability analysis, and formalizes a generalization of our previously introduced safety shield to general robotic environments.
We then present how the frequency of failsafe interventions can be reduced in~\cref{Sec_Methods}.
\cref{Sec_Experiments} discusses the main results of our experiments. 
Finally, we conclude this article in \cref{Sec_Conclusions}.

\section{PRELIMINARIES} \label{Sec_Preliminaries}
    This section introduces the required foundations of RL, set-based reachability analysis, and our safety shield.

\subsection{Reinforcement learning}
RL aims to find an optimal policy for Markov decision processes (MDPs), defined by the tuple $\left(\mathcal{S}, \mathcal{A}, R, p, \gamma\right)$~\cite{sutton_2018_ReinforcementLearning}.
This work focuses on continuous state and action spaces $\mathcal{S}$ and $\mathcal{A}$, respectively, as is common in robotic applications.
We define the state-transition probability density function $p : \mathcal{S} \times \mathcal{A} \times \mathcal{S} \rightarrow \mathopen[0, \infty\mathclose)$ to describe the probability of reaching the next state $\boldsymbol{s}_{i+1}$ when choosing action $\boldsymbol{a}_{i}$ in state $\boldsymbol{s}_{i}$.
The environment provides a reward $R : \mathcal{S} \times \mathcal{A} \times \mathcal{S} \rightarrow \mathbb{R}$ for each transition, which is discounted by the discount factor $\gamma \in \mathopen[0,1\mathclose]$.
The agent learns a stochastic policy $\pi(\boldsymbol{a}_{i} | \boldsymbol{s}_{i})$ from which action $\boldsymbol{a}_{i}$ is sampled in state $\boldsymbol{s}_{i}$.

For constrained RL, Altman~\cite{altman_1998_ConstrainedMarkov} extends the MDP by a cost function $C : \mathcal{S} \times \mathcal{A} \times \mathcal{S} \rightarrow \mathbb{R}$, a cost limit $d : \mathcal{S} \times \mathcal{A} \rightarrow\mathbb{R}$, and a cost discount $\gamma_C \in \mathopen[0,1\mathclose]$.
In practice, we limit ourselves to a fixed cost limit $d \in \mathbb{R}$. 

\subsection{Reachablility analysis}
In this work, we adopt a set-based reachability analysis approach for ensuring safety and finding low-interfering actions.
We consider systems with the dynamics $\dot{\xb}(t)=f(\xb(t), \ub(t), \wb(t))$, with bounded control inputs $\ub(t) \in \Us$, disturbances $\wb(t) \in \Ws$, and possible states $\xb(t) \in \Xs$.
We denote the control input trajectory in the time interval $[t_0, t]$ as $\ub([t_0, t])$ and adopt this notation for all signals.
Given such a control input trajectory, an initial state $\xb_0$, and a disturbance trajectory $\wb([t_0, t])$, the system follows the trajectory $\chi\left(t, \xb_0, \ub([t_0, t]), \wb([t_0, t])\right)$.
The forward reachable set $\mathcal{R}(t)$ of a system starting in the set of initial states $\Xs_0$ with unknown control inputs comprises all states reachable at time $t$:
\begin{align}
    \mathcal{R}(t) = & \left\{ \chi\left(t, \xb_0, \ub([t_0, t]), \wb([t_0, t])\right) \mid\right. \\
    & \left.\xb_0 \in \Xs_0, \forall t: \ub(t) \in \Us, \wb(t) \in \Ws \right\}\,. \nonumber
\end{align}
A general state-feedback control law $\uphix{Z}(t) = \phix{Z}(x(t)) \in \Us$ can also generate an input signal, resulting in the trajectory $\chix{Z}\left(t, \xb_0, \uphix{Z}([t_0, t]), \wb([t_0, t])\right)$.
We denote the reachable set under a feedback controller by $\RZ(t)$:
\begin{align}
    \RZ(t)=  & \left\{ \chix{Z}\left(t, \xb_0, \uphix{Z}([t_0, t]), \wb([t_0, t])\right) \mid\right. \\
    & \left.\xb_0 \in \Xs_0, \forall t: \wb(t) \in \Ws \right\}\,. \nonumber
\end{align}
All states reachable during the time interval $[t_0, t_1]$ are given by $\mathcal{R}\left(\left[t_0, t_1\right]\right)=\bigcup_{t \in\left[t_0, t_1\right]} \mathcal{R}(t)$.
We denote the set of points that a system can occupy in Euclidean space at time $t$ and a time interval by $\mathcal{O}(\mathcal{R}(t))$ and $\mathcal{O}(\mathcal{R}(\left[t_0, t_1\right]))$, respectively, and further refer to it as the reachable occupancy (RO).
We further introduce a point in Euclidean space as $\pb$, a ball as $\mathcal{B}(\mathbf{c}, r) = \left\{\pb \mid \|\pb - \mathbf{c}\|_2 \leq r\right\}$ with center $\mathbf{c}$ and radius $r$, and a function $\ball$ that overapproximates a RO with a ball $\mathcal{O} \subseteq \ball{\left(\mathcal{O}\right)} = \Bhat(\mathbf{c}, r)$.

\subsection{Safety shield}\label{sec:safety_shield}
To ensure the safety of RL agents, we utilize the safety shield for manipulators proposed in~\cite{thumm_2022_ProvablySafe} and generalize it to arbitrary robotic environments.
The safety shield relies on the existence of an ISS that is reachable from any state.
An example of an ISS is an entirely stopped robot for manipulators or mobile robots, as per ISO~10218-1~2021~\cite{iso_2021_RoboticsSafety}.
As we demonstrated in~\cite{thumm_2022_ProvablySafe}, a safety shield for RL is less restrictive when it operates on a higher frequency than the output frequency of the RL agent.
We, therefore, execute each RL action $\boldsymbol{a}_i$ for $L$ time steps and perform a safety shield update at every time step.

At each time step $k$, we calculate an intended and a failsafe trajectory.
Without loss of generality, we reset the clock to $t_0$ at each time step.
The intended trajectory $\chix{I}$ results from the desired agent action output and is executed with the control law $\phix{I}(\boldsymbol{x}(t), \boldsymbol{a})$ for $L$ time steps.
The failsafe trajectory $\chix{F}$ leads the robot to an ISS using a failsafe controller $\phix{F}(\boldsymbol{x}(t))$ in $F$ time steps, where $F$ depends on the state of the robot.
We append an entire failsafe trajectory to a single step of the intended trajectory to form a so-called shielded trajectory 
\begin{align}\label{eq:shielded_trajectory}
    \chix{S} = 
    \begin{cases}
        \chix{I}\left(t, \xb_0, \uphix{I}([t_0, t]), \wb([t_0, t])\right), &t \in [t_0, \td]\\
        \chix{F}\left(t, \xb_1, \uphix{F}([\td, t]), \wb([\td, t])\right), & t \in [\td, \tb] \, ,
    \end{cases}
\end{align}
with $D=1$, $\tb$ as the time at the end of the shielded trajectory, and $S = D + F$.

We verify the shielded trajectory by calculating the ROs of the robot $\mathcal{O}(\RS(\left[t_0, \tb\right]))$
and the $J$ obstacles $\Oobs(\left[t_0, \tb\right]) = \bigcup_{j \in J} \mathcal{O}(\Robsj(\left[t_0, \tb\right]))$ for each partial time interval in the overall time interval $[t_0, \tb]$; see \cref{fig:reachable-sets}.
We then check if the ROs are intersection-free for all partial time intervals using our open-source toolbox SaRA~\cite{schepp_2022_SaRATool}:
\begin{align}\label{eq:intersection}
     \mathcal{O}(\RS(\left[t_{k-1}, t_k\right])) \cap \Oobs(\left[t_{k-1}, t_k\right]) = \varnothing, \, \forall k \in \left\{1, \dots, S\right\}.
\end{align}
We ensure safety indefinitely through induction by assuming that the robot starts in an ISS and executing the last verified failsafe trajectory if the verification in \eqref{eq:intersection} fails.
The set-based representation of the system enables us to guarantee safety in both simulation and real-world applications, thus bridging the simulation-to-reality gap. For a detailed implementation of our safety shield for the OpenAI safety gym, we refer the reader to the Appendix.%

\section{METHODOLOGY} \label{Sec_Methods}
    \begin{figure}[t]
   	\subfloat{
   		\includegraphics[scale=1.09]{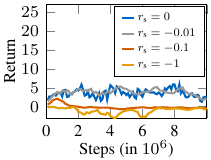}
   		\label{fig:reward_tuning}
   	}
   	\subfloat{
   		\includegraphics[scale=1.09]{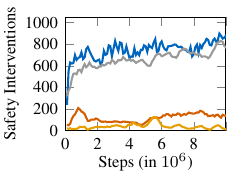}
   		\label{fig:reward_tuning_safety_activity}
   	}
    \caption{Reward and shield activation for different values of reward punishment $r_{\text{s}}$ for shield usage with the shielded PPO agent. If the reward punishment is too weak ($r_{\text{s}}=-0.01$), the shield usage is not affected, and if it is slightly too high ($r_{\text{s}}=-0.1$), the agent learns to never use the shield at all cost.}
    \label{fig:reward-shaping}
\end{figure}
This work investigates four methods for reducing the frequency of failsafe interventions.
The state-of-the-art method is to assign a fixed negative reward for each failsafe intervention~\cite{alshiekh_2018_SafeReinforcement, pham_2018_OptLayerPractical, shao_2021_ReachabilitybasedTrajectory}.
However, the evaluation of reward tuning for the \texttt{Point-Button} environment (see~\cref{Sec_Experiments}), in~\cref{fig:reward-shaping} shows that this approach often results in either no improvement in the frequency of failsafe interventions or a drastic reduction in performance.
To improve this, we introduce two safety reduction techniques, proactive replacement and projection, and compare them with constrained RL.

\subsection{Proactive replacement}
\begin{figure}[t]
    \centering
    \includegraphics[width=0.99\columnwidth]{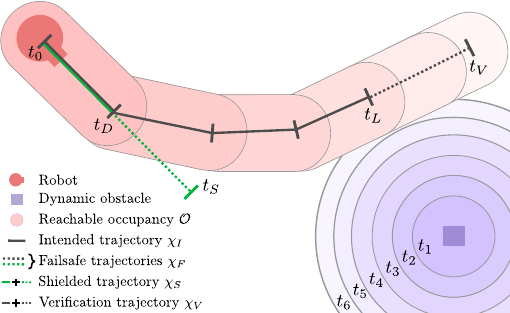}
    \caption{Four types of trajectories can be distinguished in proactive replacement: The intended trajectory results from the action output of the RL agent. To build the shielded trajectory, a failsafe trajectory is appended to a single step of the intended trajectory. The verification trajectory comprises the intended trajectory and a subsequent failsafe trajectory, and is used to validate the RL action. In this particular example, the shielded trajectory is safe, but the RL action is replaced because the verification trajectory collides with the reachable set of the dynamic obstacle.}
    \label{fig:reachable-sets}
\end{figure}

Our first approach replaces actions that result in an intersection between the ROs of the robot and obstacles.
The new action as any other action does not impair safety due to the safety shield as described in \cref{sec:safety_shield}.
To prevent interventions during the RL step, we construct a verification trajectory $\chix{V}$ as depicted in~\cref{fig:reachable-sets}, composed of the intended trajectory of the RL step followed by a failsafe trajectory.
We define the verification trajectory analogous to \eqref{eq:shielded_trajectory} except that $D$ is replaced by $L$ and $S$ is replaced by $V = L + F$.
After constructing the verification trajectory, we verify it for potential failsafe interventions by checking for intersections with the robot's ROs in the time interval $\left[t_0, \tv\right]$ using \eqref{eq:intersection}, where $S$ is replaced by $V$.

If the initial verification fails, we sample up to $M$ replacement actions uniformly from the action space and repeat the verification until an action is successfully verified.
The agent executes the first verified replacement action found during the RL step.
If we cannot find a verifiable replacement action, we execute an environment-specific neutral action, e.g., the zero vector $\boldsymbol{a} = \boldsymbol{0}$.
As a result, the proactive replacement might fail to prevent safety interventions, which we evaluate in our experiments in \cref{Sec_Experiments}.

\begin{figure}[t]
    \vspace{0.17cm} 
	\centering
	\includegraphics[width=\columnwidth]{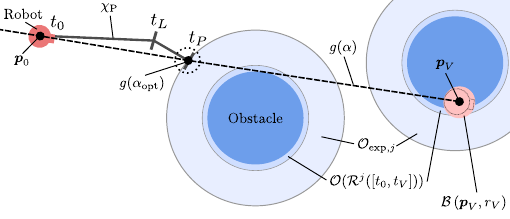}
	\caption{Our proactive projection method finds the longest intersection-free line segment from the current robot position to the end of the verification trajectory.}
	\label{fig:projection}
\end{figure}

\subsection{Proactive projection}
Our second intervention reduction approach is a type of action projection.
The goal is to find a reachable point in Euclidean space near the position of the robot at the end of the intended trajectory without colliding with the environment.
For this, we model the robot as a multi-body system and describe its RO as the union of the ROs of its $N$ bodies $\mathcal{O}\left(\mathcal{R}\right) = \cup_{n \in N} \mathcal{O}\left(\mathcal{R}^n\right)$.
We overapproximate the RO of a robot body with a capsule $\mathcal{C}\left(\boldsymbol{p}, \boldsymbol{d}, r, l\right) = \left\{\boldsymbol{p} + \beta \, l \, \frac{\boldsymbol{d}}{\|\boldsymbol{d}\|_2}\mid \beta \in \mathopen[-1, 1\mathclose] \right\} \oplus \mathcal{B} \left(\mathbf{0}, r\right)$, where $\oplus$ is the Minkowski sum\footnote{$\mathcal{A} \oplus \mathcal{B}=\{\boldsymbol{a}+\boldsymbol{b} \mid \boldsymbol{a} \in \mathcal{A}, \boldsymbol{b} \in \mathcal{B}\}$}.
The ROs of the obstacles in the time interval $\left[t_0, \tv \right]$ are expanded by the radius of the robot body capsules $r_{\text{exp}} = \rv + \epsilon$ using
\begin{align}
	\Oexp &= \bigcup_{j \in J} \Oexpj\\
    &= \bigcup_{j \in J} \mathcal{O}\left(\obsj{\mathcal{R}}(\left[t_0, \tv\right])\right) \oplus \mathcal{B}\left(\mathbf{0}, r_{\text{exp}}\right) \, . \nonumber
\end{align}
Depending on the current position $\pnull$, we select different projection strategies.
Suppose $\pnull$ is already in $\Oexp$. 
In that case, we use the projection of $\pnull$ to the nearest point outside of $\Oexp$ as the new target position
\begin{alignat}{2}\label{eq:opt_proj_0}
    &\!\min & \qquad & \tilde{\pb}^\top \tilde{\pb}\\
	&\text{subject to} & &(\pnull + \tilde{\pb}) \notin \mathcal{O}_{\text{exp}}\, , \nonumber
\end{alignat}
where $\tilde{\pb}$ is the displacement between target and current position.
To solve the non-convex optimization problem in \eqref{eq:opt_proj_0}, we use the $l_1$ penalty method for sequential convex optimization using trust regions presented \mbox{in \cite[Algorithm~1]{schulman_2014_MotionPlanning}} and refer the reader to the Appendix for implementation details.

If at the beginning of the RL step $\pnull \notin \Oexp$, we construct a straight line for each robot body part from its current Carthesian position $\pnull^n$ to its predicted Carthesian position at the end of the verification trajectory $\pv^n$ as
\begin{align}
	g^n(\alpha) &= \pnull^n + \alpha \left(\pv^n - \pnull^n\right), \quad \alpha \in \mathbb{R} \, ,
\end{align}
and \mbox{$h^n(\alpha) = \boldsymbol{d}_0^n + \alpha \left(\boldsymbol{d}_V^n - \boldsymbol{d}_0^n\right)$}.
We then look for the longest intersection-free line segment for each robot body part with $\mathcal{G}^n\left(\left[\alpha_1, \alpha_2\right]\right) = \left\{g^n(\alpha) + \beta \, l^n\, h^n(\alpha) \mid \alpha \in \mathopen[\alpha_1, \alpha_2\mathclose], \, \beta \in \mathopen[-1, 1\mathclose]\right\}$ from the current robot body part position to $\pv^n$ using
\begin{alignat}{2}\label{eq:line_segment}
	&\!\max & \qquad &\alpha_{\text{opt}} \\
	&\text{subject to} & & 0 < \alpha_{\text{opt}} < 1\, , \nonumber \\
	& & & \mathcal{G}^n\left(\left[0, \alpha_{\text{opt}}\right]\right) \cap \Oexp = \varnothing, \quad n \in 1, \, \dots, \, N \, . \nonumber
\end{alignat}
The line segment approach in \eqref{eq:line_segment} is illustrated exemplary for $h^n = \boldsymbol{0}$ and $N=1$ in~\cref{fig:projection}.
Finally, we use a trajectory planner to plan a trajectory $\chix{P}$ from $\pnull$ to an ISS in an $\epsilon$-bound around $\pv$.
The trajectory planner has to ensure that the first $L$ steps of the trajectory have a constant control input.
As the projection only checks against intersection in the end-configuration of the verification trajectory, it is not guaranteed that the resulting projected trajectory is intersection-free in \eqref{eq:intersection} with $S$ being replaced by $V$. 
Therefore, if $\chix{P}$ would result in an intersection, we can repetitively reduce $\alpha$ for $M$ times to find a more conservative projection point.
If we cannot project the action to a verifiable trajectory $\chix{P}$, we execute the neutral action and continue to ensure safety with our safety shield.

\subsection{Constrained RL}
We compare the two presented methods with a constrained RL approach that aims to minimize the number of failsafe interventions in an episode. 
We assign a constant cost $C_F$ to each failsafe intervention in an RL step and train a PID-Lagrangian agent to perform the environment task while staying below a threshold of safety interventions.
This approach minimizes safety interventions without any additional implementation effort.

\section{EXPERIMENTS} \label{Sec_Experiments}
    This section discusses the effects of our proposed intervention reduction methods in two different applications, the OpenAI safety gym benchmark and a human-robot collaboration setting.
\subsection{OpenAI safety gym}
\begin{figure*}[t]
    \vspace{0.17cm} 
	\centering
	\subfloat{
		\includegraphics[scale=1]{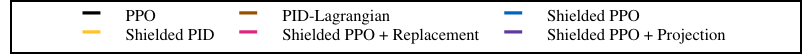}
	}\\[-0.3cm]
    \subfloat{
    \includegraphics[scale=1]{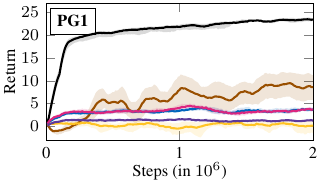}
    \label{fig:reward_pg1}
    }
    \subfloat{
    \includegraphics[scale=1]{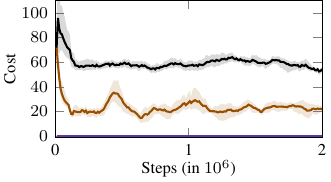}
    \label{fig:cost_pg1}
    }
    \subfloat{
    \includegraphics[scale=1]{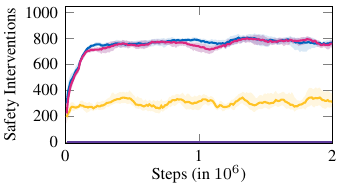}
    \label{fig:safety_pg1}
    }\\[-0.3cm]
	\subfloat{
		\includegraphics[scale=1]{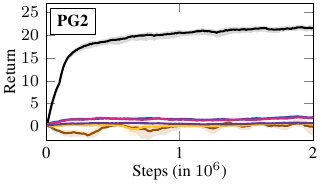}
		\label{fig:reward_pg2}
	}
	\subfloat{
		\includegraphics[scale=1]{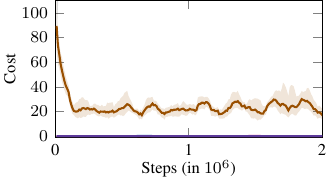}
		\label{fig:cost_pg2}
	}
	\subfloat{
		\includegraphics[scale=1]{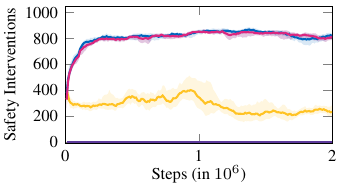}
		\label{fig:safety_pg2}
	}\\[-0.3cm]
	\subfloat{
		\includegraphics[scale=1]{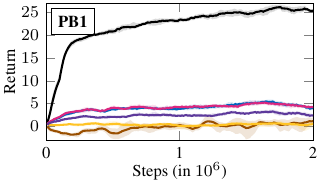}
		\label{fig:reward_pb1}
	}
	\subfloat{
		\includegraphics[scale=1]{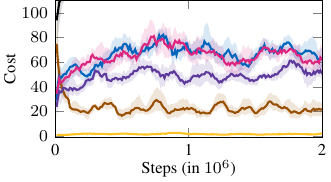}
		\label{fig:cost_pb1}
	}
	\subfloat{
		\includegraphics[scale=1]{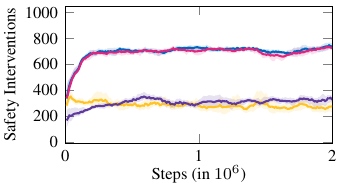}
		\label{fig:safety_pb1}
	}\\[-0.3cm]
	\subfloat{
		\includegraphics[scale=1]{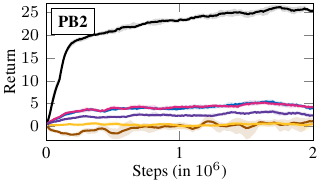}
		\label{fig:reward_pb2}
	}
	\subfloat{
		\includegraphics[scale=1]{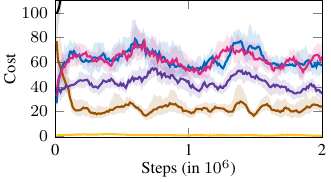}
		\label{fig:cost_pb2}
	}
	\subfloat{
		\includegraphics[scale=1]{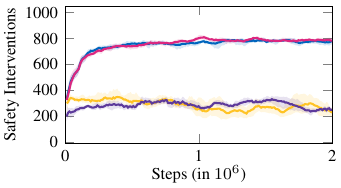}
		\label{fig:safety_pb2}
	}
	\caption{Mean and its \num{95}\% confidence interval over ten random seeds for the OpenAI safety gym environments. From top to bottom: \texttt{Point-Goal1}, \texttt{Point-Goal2}, \texttt{Point-Button1}, and \texttt{Point-Button2}. From left to right: the return, cost, and number of safety interventions per episode with length \num{1000}.
	Our proactive projection method reduces the number of failsafe interventions drastically with only slight performance loss.
	The costs for the shielded agents in the Button environments are non-zero because we did not account for incorrectly pushed buttons. There are no other collisions with the static environment or the gremlins.
	}
	\label{fig:results_safety_gym}
\end{figure*}
Our OpenAI safety gym experiments evaluate the proposed safety reduction methods on two continuous control tasks: \texttt{Point-Goal} and \texttt{Point-Button}.
For both tasks, we consider a cylindrical robot modeled as a point mass that is controlled by its yaw rate and acceleration in the direction of travel. 
The agent perceives its velocity, acceleration, and distance to the target and the obstacles through a velocity, an acceleration, and a LiDAR sensor, respectively.
Each episode starts with randomized positions of the robot, obstacles, and goals.

The \texttt{Point-Goal} task requires the agent to navigate to a target area while avoiding hazards.
On the other hand, in the \texttt{Point-Button} task, the agent needs to move to the correct button from multiple options while avoiding dynamic obstacles called gremlins.
For both tasks, the agent incurs a cost of \num{1} for being in a hazard or an incorrect button area and for colliding with a vase or a gremlin.
The shielded PID-Lagrangian agent additionally receives a cost of $C_F = 1$ for failsafe interventions.
Both tasks have two difficulty levels, with level \num{2} featuring more obstacles.
Each episode lasts \num{1000} RL steps, with a new goal randomly selected upon completion.
All training results in \cref{fig:results_safety_gym} show the mean metrics and their \num{95}\% confidence intervals\footnote{We use the default values of the \texttt{scipy} bootstrap function.} over \num{2} million RL steps on ten random seeds.
We kept the training hyperparameters from~\cite{stooke_2020_ResponsiveSafety} for all training runs on every environment.

In the \texttt{Point-Button} environments, we exclude button constraints from our safety shield as buttons trigger instantaneously from being eligible to invalid upon contact.
This breaks the assumption that a stopped robot is in an ISS.
Therefore, we only test the capabilities of our shielded PID-Lagrangian agent to reduce shield interventions and incorrect button violations simultaneously.

Our results in~\cref{fig:results_safety_gym} demonstrate the effectiveness of our safety shield in mitigating all environmental costs in the \texttt{Point-Goal} tasks and all costs except those due to button constraints in the \texttt{Point-Button} task, serving as the first provably safe benchmark in the OpenAI safety gym.
Our proactive projection method significantly reduces failsafe interventions of the shielded PPO agent across all environments.
In contrast, the proactive replacement method is less effective in lowering safety interventions due to its failure to find a verified replacement action when using our randomized selection.
The proactive projection method receives slightly less return than our shielded PPO agent, but still performs better than the shielded PID-Lagrangian agent.
In the \texttt{Point-Button} tasks, the PID-Lagrangian agent reduces the incorrect button presses significantly.
This indicates that constrained RL is a suitable technique for reducing non-safety-critical constraints, which would be hard to shield against otherwise.

\subsection{Human-robot collaboration}
\begin{figure}[t]
    \vspace{0.17cm} 
	\centering
	\subfloat{
		\includegraphics[scale=0.8]{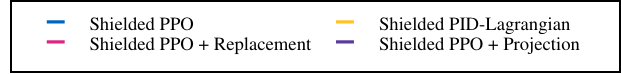}
	}\\[-0.3cm]
    \subfloat{
    	\includegraphics[scale=1.05]{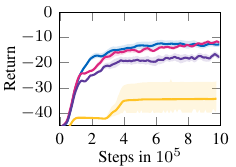}
    	\label{fig:reward_hrc}
    }
    \subfloat{
    	\includegraphics[scale=1.05]{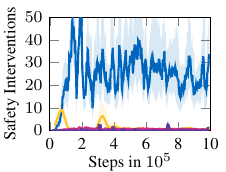}
    	\label{fig:safety_activity_hrc}
    }
    \caption{Mean and its \num{95}\% confidence interval over five random seeds of the cumulative reward and failsafe interventions per episode for the human-robot collaboration environment with 1 million RL steps and \num{400} RL steps per episode.}
    \label{fig:human-robot-results}
\end{figure}

To test the transferability of our findings to more complex robotics tasks, we trained a reaching task in a human-robot simulation. 
We deployed the trained agents on a Schunk LWA 4P manipulator in a real-world setting.
Our six DoF manipulator has to move from a given initial end effector position $\pb_{\text{init}}$ to a target end effector position $\pb_{\text{goal}}$, with the episode terminating upon goal achievement.
The RL agent receives two types of observations: the difference between the target and current end effector position in all three dimensions and the Euclidean distance between the end effector and the human head, i.e., $\boldsymbol{s} = [(\pb_{\text{goal}}-\pnull)^\top, \|\pnull-\pb_{\text{head}}\|_2]^\top$.
The RL actions are the desired change in end effector positions from the current position $\pnull$ to the next position $\pL$, i.e., $\pL = \pnull + \boldsymbol{a}$.
Our proactive projection technique is only applied to the RO of the end effector, i.e., $N=1$, to reduce calculation time. 
The human in the environment follows pre-recorded movements with randomized start positions.
As safety is paramount in human environments, we only compare the shielded agents and refer to~\cite{thumm_2022_ProvablySafe} for a detailed comparison with unsafe agents.

\cref{fig:human-robot-results} shows that the shielded PPO agent performs well in the reaching task but triggers a failsafe intervention in \num{2.5}$\%$ of the RL steps, with up to \num{7.5}$\%$ at the beginning of the training.
Our proposed failsafe prevention techniques reduce the number of safety interventions by a factor of \num{10} to around \num{0.25}$\%$, with no decrease in performance for proactive replacement and only a slight decrease for the projection method.
Contrary to the OpenAI safety gym environment, the PID-Lagrangian agent is unable to learn a suitable policy that is both high-performing and reduces failsafe interventions, even after exhaustive hyperparameter optimization.
We attribute the limited success of PID-Lagrangian to the increased complexity of the human-robot environment.

\begin{figure}
    \vspace{0.17cm} 
	\centering
	\includegraphics[width=0.8\columnwidth]{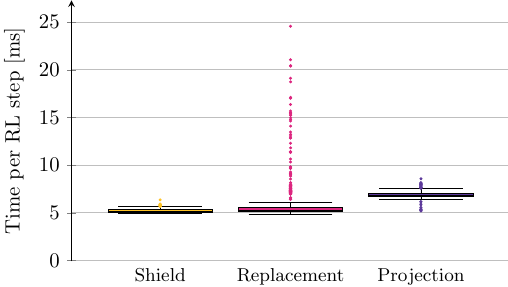}
	\caption{Comparison of calculation times per RL step for the shielded agent, the shielded agent plus proactive replacement, and the shielded agent plus proactive projection in the human-robot environment. We report the median (black line), \num{25}\%- and \num{75}\%-quantile (box), the smallest value that is larger than the \num{25}\%-quantile $- \ 1.5 \  \cdot$ interquartile range (lower whisker), the largest value that is smaller than the \num{75}\%-quantile $+ \ 1.5 \ \cdot$ interquartile range (upper whisker), and outliers (dots).}
	\label{fig:runtime_comparison}
\end{figure}

The runtime analysis\footnote{Run on an Intel\textsuperscript{\textregistered} Core\texttrademark~i7-10750H CPU @ 2.60GHz × 12 and 32GB DDR4 RAM @ 3.2 GHz.} in \cref{fig:runtime_comparison} illustrates that both proactive replacement and proactive projection do not add a significant overhead to the calculation time in each RL step, despite the highly complex environment.
Proactive replacement shows significant outliers that occur when no replacement action can be found.
These outliers can be reduced by lowering the number of resamples, or selecting a different replacement strategy.

We also present the performance of our agents on a real-world robot in our supplementary video.
The shielded PPO agent consistently fulfills the task, while the action replacement and projection techniques allow the agent to move away from the human RO, resulting in smoother trajectories and fewer safety interventions.

\section{CONCLUSIONS} \label{Sec_Conclusions}
    Our results clearly show that our proposed proactive projection method significantly reduces the frequency of failsafe interventions while maintaining competitive performance.
Contrary to existing intervention reduction methods, our approach does not require careful parameter tuning to reduce failsafe interventions, making it easily applicable in complex robotic environments.
The results of evaluations on the real manipulator further strengthen our findings and make us confident that our approach is transferable to other tasks.
To further demonstrate the effectiveness of combining a safety shield with proactive projection, we intend to test it in more challenging real-world human-robot collaboration environments, such as construction sites or manufacturing facilities.
These environments pose additional safety challenges due to heavy machinery and high-risk tasks and thus provide an opportunity to evaluate the robustness of our approach.

\section*{APPENDIX}\label{Sec_Appendix}
    \subsection{Adaptions to safety shield for point robot}
In the OpenAI safety gym, we consider a robot modelled as a point mass, whose state can be described by its position $\pb^\top=\left[p_\xm, p_\ym\right]$, velocity $\vb^\top=\left[v_\xm, v_\ym\right]$, and orientation $\varphi$, i.e., $\xb = \left[p_\xm, p_\ym, v_\xm, v_\ym, \varphi\right]^\top$. The robot has two inputs $\boldsymbol{u} = \left[u_1, u_2\right]$. The nonlinear system dynamics are given by~\cite{ray_2019_BenchmarkingSafe}
\begin{subequations}
\label{eq:point_dynamics}
\begin{align}
    \dot{\pb} &= \vb,\label{eq:point_dynamics3}\\
    \dot{\vb} &= \mathbf{R}_\zm(\varphi) \left[u_1, 0\right]^\top - \vb \frac{k_d}{m}\label{eq:point_dynamics2}\\
    \dot{\varphi} &= u_2\, ,\label{eq:point_dynamics1}
\end{align}
\end{subequations}
where $\mathbf{R}_\zm(\varphi)$ describes the rotation matrix around the $\zm$-axis with angle $\varphi$, $k_d = \SI{0.01}{\kilogram \per \second}$,
and the mass of the robot is $m = \SI{5.19}{\gram}$. 
The input is limited by $|u_1| \leq \uamax = \SI{9.63}{\meter \per \second \squared}$ and $|u_2| \leq \ubmax = \SI{1}{\per \second}$.
It is important to note that the robot dynamics are only approximated by these equations and do not necessarily reflect the physical behavior of a real-world robot.

We directly obtain the intended trajectory  from \eqref{eq:point_dynamics}.
The aim of the failsafe trajectory is to brake the robot as fast as possible, so the objective function is
\begin{align}
    \max_{u_1, u_2}\left(\|\vb\|_2^2 - \|\vb + \dt \dot{\vb}\|_2^2\right)\, ,
\end{align}
which has a global optimum at $u_{1, \text{opt}} = (\cos(\varphi) v_\xm + \sin(\varphi) v_\ym) (\frac{k_d}{m} - \frac{1}{\dt})$ and $u_{2, \text{opt}} = (\atan2(v_\ym, v_\xm) - \varphi) / \dt$.
Therefore, the failsafe controller uses $\boldsymbol{u} = \left[\clip\left(u_{1, \text{opt}}, -u_{1, \text{max}}, u_{1, \text{max}}\right), \clip\left(u_{2, \text{opt}}, -u_{2, \text{max}}, u_{2, \text{max}}\right)\right]$, where $\clip(a, b, c)$ clips $a$ to the interval $[b, c]$.

For the reachability analysis, we consider the \mbox{OpenAI} safety gym as a two-dimensional environment.
The reachable occupancies of the obstacles are calculated using the velocity-constrained model proposed by Liu et al. \cite{liu_2017_ProvablySafe}.
To calculate the reachable occupancy of the point robot, only the position is relevant, so we define its trajectory with abuse of notation as the solution of \eqref{eq:point_dynamics3} $\chix{\pb}\left(t, \xb_0, \ub([t_0, t]), \wb([t_0, t])\right)$.
To achieve fast calculation times, we linearly interpolate the trajectory of the robot between two shield steps $t_0$ and $t_1$ as 
\begin{align}\label{eq:linearization}
    \tilde{\chi}_{\pb}\left(t, \xb_0, \ub([t_0, t]), \wb([t_0, t])\right) = \xi \pb(t_0) + (1-\xi) \pb(t_1) \, ,
\end{align}
with $\xi \in \left[0, 1\right]$.
The resulting linearization error in the position can be over-approximated by
\begin{align}
    \zeta = \left\|\tilde{\chi}_{\pb}\left(\frac{t_1+t_0}{2}\right)-\chi_{\pb}\left(\frac{t_1+t_0}{2}\right)\right\|_2 \leq \frac{d^2 p(t)}{d t^2} \frac{(\dt)^2}{8} \, ,
\end{align}
as shown by Beckert et al. \cite{beckert_2017_OnlineVerification} for general point movements with known acceleration limits.
For the point robot, this results in $\zeta = \frac{a_{0, \text{max}} (\dt)^2}{8 m}$.
The reachable set of the position of the robot is therefore 
\begin{align}
    \mathcal{R}_{\pb}(t) = (\xi(t) \pb(t_0) + (1-\xi(t)) \pb(t_1)) \oplus \mathcal{B}(\mathbf{0}, \zeta) \, ,
\end{align}
where $\xi(t) = \frac{t_1 - t}{t_1 - t_0}$.
To get the reachable occupancy of the robot, we simply add its radius $r$ to the reachable set
\begin{align}
    \mathcal{O}(\mathcal{R}_{\pb}(t)) = \mathcal{R}_{\pb}(t) \oplus \mathcal{B}(\mathbf{0}, r) \, .
\end{align}
The linearization in \eqref{eq:linearization} simplifies the reachable occupancy of the robot between two shield steps to a capsule, which allows for fast intersection checking with the environment.

The OpenAI safety gym only provides lidar measurements of the obstacles, which can result in occlusions of obstacles by other obstacles.
Such occlusions can be handled using set-based predictions as shown exemplary for an autonomous driving task in~\cite{orzechowski_2018_TacklingOcclusions}. 
However, this solution increases the size of the reachable sets, making the safety shield more restrictive. 
To mitigate this issue, we decided to augment the safety shield with precise information about the exact position of each obstacle.
The shielded RL agent, however, does not have access to the exact positions, ensuring a fair comparison with unshielded agents.

\subsection{Implementation details for proactive projection}
We specify the constraint function of \eqref{eq:opt_proj_0} as 
\begin{alignat}{2}\label{eq:opt_proj_sd}
    -\sd\left(\pnull + \tilde{\pb}, \Oexpj\right) \leq 0, \forall j \in J \, .
\end{alignat}
Here, $\sd\left(\pb, \mathcal{O} \right)$ is the signed distance between a point $\pb$ and a set $\mathcal{O}$ with
\begin{align}
    \sd\left(\pb, \mathcal{O}\right) = \sign\left((\pb_E - \pb)^\top (\pb_C - \pb)\right) \|\pb_E - \pb\|_2 \, ,
\end{align}
where $\pb_E$ is the closest point on the edge of $\mathcal{O}$ to $\pb$, and $\pb_C$ is the center of $\mathcal{O}$.
As the projection at the beginning of an RL step is time-critical, we use a simplified linearization of the inequality constraints for the \textit{convexify} step of \cite[Algorithm~1]{schulman_2014_MotionPlanning}:
\begin{align}
    h^j(\tilde{\pb}) &= - \sd\left(\pnull + \tilde{\pb}, \Oexpj\right)\\
    &\approx \sd\left(\pnull, \Oexpj\right) \frac{\pb^j_E - \pnull}{\|\pb^j_E - \pnull\|_2^2} \tilde{\pb} - \sd\left(\pnull, \Oexpj\right) \, .
\end{align}

\section*{ACKNOWLEDGMENT}
The authors gratefully acknowledge financial support by
the Horizon 2020 EU Framework Project CONCERT under grant 101016007.

\addtolength{\textheight}{-6.0cm}
\bibliographystyle{IEEEtran}
\bibliography{library}

\end{document}